\documentclass{llncs}

\usepackage{newlfont}
\usepackage[latin1]{inputenc}
\usepackage{amsmath}
\usepackage{amsfonts}
\usepackage{url}
\usepackage{amssymb}
\usepackage{verbatim}
\usepackage{graphicx}
\begin{document}
\newcommand{\RR}{\mbox{\bf R}}

\mainmatter              
\title{On Clustering Time Series Using Euclidean Distance and Pearson Correlation}
\titlerunning{On Clustering Time Series}
\author{Michael R.\ Berthold \and Frank H\"{o}ppner}
\authorrunning{M.R.\ Berthold, F.\ H\"{o}ppner}
\tocauthor{}
\institute{
  Dept.\ of Computer and Information Science,
  University of Konstanz,
  Germany\\
  \email{Michael.Berthold@uni-konstanz.de}
  \and
  Dept.\ of Economics,
  Univ.\ of Applied Sciences Wolfsburg,
  Germany\\
  \email{F.Hoeppner@fh-wolfsburg.de}
}

\maketitle

\begin{abstract}
For time series comparisons, it has often been observed that z-score
normalized Euclidean distances far outperform the unnormalized
variant. In this paper we show that a z-score normalized, squared
Euclidean Distance is, in fact, equal to a distance based on Pearson
Correlation. This has profound impact on many distance-based
classification or clustering methods. In addition to this
theoretically sound result we also show that the often used k-Means
algorithm formally needs a modification to keep the interpretation as
Pearson correlation strictly valid. Experimental results demonstrate
that in many cases the standard k-Means algorithm generally produces
the same results.
\end{abstract}

\section{Introduction}

In the KD-Nuggets poll of July 2007 for the most frequently analyzed
type of data, time series data was voted on the second place. It is
therefore not surprising, that a large number of papers with
algorithms to cluster, classify, segment or index time series have
been published in recent years. For all these tasks, measures to
compare time series are needed, and quite a number of measures have
been proposed (cf.~\cite{Keogh:DMKD:2004}). Among these measures, the
Euclidean distance is by far the most frequently used distance measure
(although many more sophisticated measures
exist)~\cite{Keogh:DMKD:2004} with applications in finance
\cite{Gavrilow:KDD:2000}, medicine \cite{Caraca-Valente:KDD:2000} or
electric power load-forecast \cite{Ruzic:TPS:2003}, to mention only a
few. However, many authors then realize that normalizing the Euclidean
distance before applying standard machine learning algorithms hugely
improves their results. 

In this paper, we closer investigate the popular combination of
clustering time series data together with a normalized Euclidean
distance. The most commonly used learning method for clustering tasks
is the k-Means algorithm \cite{McQueen:SMSP:1967}. We show that a
z-score normalized squared Euclidean distance is actually equal (up to
a constant factor) to a distance based on the Pearson correlation
coefficient. This oberservation allows for standard learning methods
based on a Euclidean distance to use a Pearson correlation coefficient
by simply performing an appropriate normalization of the input
data. This finding applies to all learning methods which are based
solely on a distance metric and do not perform subsequent operations,
such as a weighted average. Therefore reasonably simple methods such as
k Nearest Neighbor and hierarchical clustering but also more complex
learning algorithms based on e.g.\ kernels can be converted into using
a Pearson correlation coefficient without actually touching the
underlying algorithm itself.

In order to show how this affects algorithms that do perform
additional e.g.\ averaging operations, we chose the often used k-Means
algorithm which requires a modification of the underlying clustering
procedure to work correctly.  Experiments indicate, however, that the
difference between the standard Euclidean-based k-Means and the
Pearson-based k-Means do not have a high impact on the results,
therefore indicating that even in such cases a simple data
preprocessing can turn an algorithm using a Euclidean distance into
its equivalent based on a Pearson correlation coefficient.

Our paper is organized as follows. In section~\ref{distance:sec} we
briefly recap distances for time series. Afterwards we show that a
properly normalized Euclidean distance is equivalent to a distance
based on the Pearson correlation coefficient. After this theoretically
sound result we quickly review k-means and the required modifications
to operate on a Pearson correlation coefficient.  In
section~\ref{sec:experiments} we perform experiments that demonstrate
the (small) difference in behaviour for standard k-Means with
preprocessed inputs and the modified version.

\section{Measuring Time Series Similarity}\label{distance:sec}

In this section, we briefly review the Euclidean distance which is
frequently used by data miners and a distance based on Pearson
Correlation which is more often used to measure the strength and
direction of a linear dependency between time series. Suppose we
have two time series $r$ and $s$, consisting of $T$ samples each
$r=(r_{1}, r_{2}, ..., r_{T})\in \RR^{T}$.

\paragraph{Euclidean Distance.}

The {\em squared Euclidean distance} between two time series $r$ and
$s$ is given by:
\begin{equation}
d_E(r,s) = \sum_{t=1}^T (r_t-s_t)^2
\label{eq:euclidean}
\end{equation}
The Euclidean distance is a metric, that is, if $r$ and $s$ have zero
distance, then $r=s$ holds. For time series analysis, it is often
recommended to normalize the time series either globally or locally to
tolerate vastly different ranges \cite{Keogh:DMKD:2004}.

\paragraph{Pearson Correlation Coefficient.}

The Pearson correlation coefficient measures the correlation
$\varrho$ between two random variables $X$ and $Y$:
\begin{equation}\label{pearsoncorr:eq}
 \varrho_{X,Y}
= \frac{ E[ (X-\mu_X)(Y-\mu_Y) ] }{ \sigma_X \sigma_Y } 
\end{equation}
where $\mu_X$ denotes the mean and $\sigma_X$ the standard
deviation of $X$. We obtain a value of $+/- 1$ if $X$ and $Y$ are
perfectly (anti-) correlated and a value of $\approx 0$ if they are
uncorrelated.

In order to use the Pearson correlation coefficient as a distance
measure for time series it is desirable to generate low distance
values for positively correlated (and thus similar) series. The {\em
Pearson distance} is therefore defined as
\begin{equation}\label{pearsondist:eq}
d_P(r,s) = 1 - \varrho_{r,s}
 =  1 - \frac{\frac{1}{T}\sum_{t=1}^T (r_t-\mu_r)(s_t-\mu_s)}{\sigma_r\sigma_s}
\end{equation}
such that $0 \le d_P(r,s) \le 2$. We obtain a perfect match (zero
distance) for time series $r$ and $s$ if there is an
$\alpha,\beta\in\RR$ with $\beta>0$ such that $r_i= \alpha +\beta s_i$
for all $1\le i\le T$. Thus, a clustering algorithm using this
distance measure is invariant with respect to shift and scaling of the
time series.

\section{Normalized Euclidean Distance vs.\ Pearson Coefficient}\label{normalized:sec}

In this section we show that a squared Euclidean Distance can be
expressed by a Pearson Coefficient as long as the Euclidean Distance
is normalized appropriately (to zero mean and unit variance). Note
that this transformation is applied and recommended by many authors
as noted in \cite{Keogh:DMKD:2004}.

We consider the squared Euclidean distance as given in
equation~\ref{eq:euclidean}:
\begin{eqnarray*}
d_E(r,s) & = & \sum_{t=1}^T (r_t-s_t)^2\\
   &=& \sum_{t=1}^T r_t^2 - 2\sum_{t=1}^T r_t s_t + \sum_{t=1}^T s_t^2\\
   &=& \underbrace{\sum_{t=1}^T (r_t-0)^2}_{(a)}
       - 2\sum_{t=1}^T r_t s_t 
       + \underbrace{\sum_{t=1}^T (s_t-0)^2}_{(b)}\\
\end{eqnarray*}
the first resp.\ third part (a/b) of this equation reflect $T$ times the
standard deviation of series $r$ resp.\ $s$ assuming that the series
are normalized to have a mean $\mu_{r/s}=0$. Assuming that the normalization
also ensures both standard deviations be $\sigma_{r/s}=1$ (resulting in
terms (a) and (b) to be equal to $T$), we can
simplify the above equation to:
\begin{eqnarray*}
d_E(r_{\mbox{\footnotesize norm}},s_{\mbox{\footnotesize norm}})
   & = & 2\cdot T - 2\sum_{t=1}^T r_{\mbox{\footnotesize norm},t}\cdot s_{\mbox{\footnotesize norm},t}\\
   & = & 2\cdot T\left( 1 - 
     \frac{\frac{1}{T}\sum_{t=1}^T (r_{\mbox{\footnotesize norm},t}-0)(s_{\mbox{\footnotesize norm},t}-0)}{1\cdot 1}
                 \right)\\
  & = & 2\cdot T\cdot d_P(r_{\mbox{\footnotesize norm}},s_{\mbox{\footnotesize norm}})
\end{eqnarray*}
Therefore the Euclidean distance of two normalized series is exactly
equal to the Pearson Coefficient, bare a constant factor of $2T$.

The equivalence of normalized Euclidean distance and Pearson
Coefficient is particularly interesting, since many published results
on using Euclidean distance functions for time series similarities
come to the finding that a normalization of the original time series
is {\em crucial}. As the above shows, these authors may in fact end up
simulating a Pearson correlation coefficient or a multiple of it.

This result can be applied to any learning algorithm which relies
solely on the distance measures but does not perform additional
computations based on the distance measure, such as averaging or
other aggregation operations. This is the case for many simple algorithms
such as k Nearest Neighbor and hierarchical clustering but also applies
to more sophisticated learning methods such as kernel methods.

However, some algorithms, such as the prominent k-Means algorithm do
perform a subsequent (sometimes weighted) averaging of the training
instances to compute new cluster centers. Here we can not simply apply
the above and -- by normalizing the input data -- introduce a different
underlying distance measure. However, since quite often only a ranking
of patterns with respect to the underlying distance function is
actually used it would be interesting to see if and how this
observation can be used to apply also to such learning algorithms
without actually adjusting the entire algorithm. In the following
section we will show how this works for the well known and often used
$k$-Means Clustering algorithm.

\section{Brief Review of $k$-Means}

The $k$-Means algorithm~\cite{McQueen:SMSP:1967} is one of the most
popular clustering algorithms. It encodes a partition into $k$
clusters by means of $k$ prototypical data points $p_i$, $1\le i\le
k$, and assigns every record $x_j$, $1\le j\le n$, to its closest
prototype. It minimizes the objective function
\begin{equation}\label{J:eq}
J = \sum_{j=1}^n \sum_{i=1}^c u_{i,j}\cdot d_{i,j}
\end{equation}
where $u_{i,j}\in\{0,1\}$ encodes the partition ($u_{i,j}=1$ iff
record $j$ is assigned to cluster/prototype $i$) and $d_{i,j}$ is the
distance between record $j$ and prototype $i$. The classical $k$-Means
uses the squared Euclidean distance as $d_{i,j}=d_E(x_j,p_i)$.

In the batch version of $k$-Means the prototypes and cluster
memberships are updated alternatingly to minimize the objective
function (\ref{J:eq}). In each step, one part of the parameters
(cluster memberships resp.\ cluster centers) is kept fixed while the
other is optimized. Hence the two phases of the algorithm first assume
that the prototypes are optimal and determine how we can then minimize
$J$ wrt the membership degrees. Since we allow values of 0 and 1 only
for $u_{i,j}$, we minimize $J$ by assigning a record to the closest
prototype (with the smallest distance):
\begin{equation}\label{uupd:eq}
u_{i,j} = \left\{ \begin{array}{ll}
 1 & \mbox{if~} d_{i,j}=\min_k d_{k,j}\\
 0 & \mbox{otherwise}\\
\end{array}\right.
\end{equation}
Afterwards, assuming that the memberships are optimal, we then find
the optimal position of the prototypes. The answer for this question
can be obtained by solving the (partial) derivatives of $\partial
J/\partial p_i = 0$ for $p_i$ (necessary condition for $J$ having a
minimum). In case of an Euclidean distance, the optimal position is
then the mean of all data objects assigned to cluster $p_i$:
\begin{equation}\label{mean:eq}
 p_i = \frac{\sum_{j=1}^n \sum_{i=1}^c u_{i,j} x_j}{\sum_{i=1}^c u_{i,j}}
\end{equation}
where the denominator represents the number of data objects assigned
to prototype $p_i$. The resulting $k$-Means algorithm is shown below:

\begin{center}
\begin{tabular}{l}
initialize prototypes (i.e.\ randomly draw from input data)\\
repeat\\
~~~update memberships using eq.~(\ref{uupd:eq})\\
~~~update prototypes using eq.~(\ref{mean:eq})\\
until convergence\\
\end{tabular}
\end{center}

\section{k-Means Clustering using Pearson Correlation}\label{clustering:sec}

The $k$-Means algorithm has its name from the prototype update
(\ref{mean:eq}), which re-adjusts each prototype to the mean of its
assigned data objects. The reason why the mean delivers the optimal
prototypes is due to the use of the Euclidean distance measure -- if
we change the distance measure in (\ref{J:eq}), we have to check
carefully if the update equations need to be changed, too. We have seen in
section~\ref{normalized:sec} that the Euclidean distances becomes the
Pearson Coefficient (up to a constant factor) if the vectors are
normalized. But if the distances are actually identical, should
not the prototype update also be the same?

The equivalence $2 d_E(r,s)=d_P(r,s)$ holds only for {\em normalized}
time series $r$ and $s$. To avoid constantly normalizing time series
in the Pearson Coefficient (\ref{pearsoncorr:eq}), we transform all
series during preprocessing, that is, we replace $r_i=(r_{i,t})_{1\le
t \le T}$ by $\hat{r}_i=(\hat{r}_{i,t})_{1\le t \le T}$ with
$\hat{r}_{i,t} = \frac{r_{i,t}-\mu_r}{\sigma_r}$, $1\le t \le T, 1\le
i\le n$. But in the k-Means objective function (\ref{J:eq}) the
arguments of the distance function are always pairs of data object and
prototype. While our time series data has been normalized at the
beginning, this does not necessarily hold for the prototypes.

If the prototypes are calculated as the mean of normalized time series
(standard k-Means), the prototypes will also have zero mean, but a
unit variance can not be guaranteed. Once the prototypes have been
calculated, their variance is no longer 1 and the equivalence no
longer holds. Therefore, in order to stick to the Pearson Coefficient,
we must include an additional constraint on the prototypes while
minimizing (\ref{J:eq}).

We use Lagrange multipliers $\lambda_i$, one for each cluster, to
incorporate this constraint in the objective function. So we arrive
at:
\begin{equation}
 J_P = \sum_{i=1}^n \sum_{j=1}^c u_{i,j} (1-x_j^\top p_i) + \sum_{i=1}^n \lambda_i (\|p_i\|^2-1) 
\end{equation}
As with classical k-Means we arrive at the new prototype update
equations by solving $\partial J/\partial p_i = 0$ for the prototype
$p_i$ (see appendix for the derivation):
\begin{equation}\label{pupd:eq}
p_{i,t} = \frac{1}{ \sqrt{ \sum_{t=1}^T \left( \sum_{j=1}^n u_{i,j} x_{j,t} \right)^2 } } \sum_{j=1}^n u_{i,j} x_{j,t}
\end{equation}
This corresponds to a normalization of the usually obtained prototypes
to guarantee the constraint of unit variance. Note that formally, due
to a division by zero, it is not possible to include constant time
series when using Pearson distance (because normalization leads to
$x_{j,t}=0$ for all $t$ and we would have a division by zero).

We now have a version of k-Means which uses the Pearson correlation
coefficient as underlying distance measure. But does the difference
between the different update equations~\ref{pupd:eq} and~\ref{mean:eq}
really matter? In the following we will show via a number of experiments
that the influence of this optimization is small.

\section{Experimental Evaluation}\label{sec:experiments}

The goal of the experimental evaluation is twofold. The k-Means
algorithm has been used frequently with z-score normalized time series
in the literature. We have seen that in this case a semantically sound
solution would require the modifications of section
\ref{clustering:sec}. Do we obtain {\em ``wrong''} results when using
standard k-Means? Will the corrected version with normalization yield
different results at all? And if so, how much will the resulting
cluster differ from each other on the kind of data used typically in
the literature.

\subsection{Artificial Data: Playing Devils Advocate}

We expect that the influence of the additional normalization step on
the results is rather small. An unnormalized prototype $p_i$ in
k-Means is now normalized by an additonal factor of $\alpha_i =
1/\|p_i\|$ in the modified k-Means. The distances of all data objects
to this prototype miss this factor. However, if the factors $\alpha_i$
are approximately the same for all prototoypes $p_i$, we make the same
relative error with all clusters and the rank of a cluster with
respect to the distance does not change. Since $k$-Means assigns each
data object to the top-ranked cluster, the differences in the
distances then do not influence the prototype assignment. Note also
that this (presumably small) error does not propagate further, since
we re-compute the average again from the normalized input data during
each iteration of the algorithm.

Before presenting some experimental results, let us first discuss
under what circumstances a difference between both versions of k-Means
may occur. Let us assume for the sake of simplicity that a cluster
consists of two time series $r$ and $s$ only. Then the prototype of
standard k-Means is $p=\frac{r+s}{2}$. Since the mean of $p$ will then
also be zero, the variance of the new prototype $p$ is identical to
its squared Euclidean norm $\|p-\bar{p}\|^2=\|p-0\|^2=\|p\|^2$, so we
have:
\begin{eqnarray*}
\|p\|^2 = \left\|\frac{r+s}{2}\right\|^2 = \frac{1}{4} \left( \|r\|^2+ \|s\|^2+ 2r^\top s\right)
\end{eqnarray*}
From the Cauchy Schwarz inequality and the z-score normalization of
$r$ and $s$ (in particular $\|r\|=\|s\|=1$) we conclude $r^\top
s\in[-1,1]$ since $(r^\top s)^2 \le \|r\|^2\cdot\|s\|^2 = 1\cdot
1$. Apparently $\|p\|$ becomes maximal, if $r^\top s$ or equivalently
the Pearson Coefficient becomes maximal. The new prototype $p$ will
have unit variance if and only if $r$ and $s$ correlate perfectly
(Pearson Coefficient of 1). If the new prototype $p$ is build from
data that is {\em not} perfectly correlated, its norm $\|p\|$ will be
smaller than one -- the worse the correlation, the smaller the norm of
$p$. As a consequence, for any subsequent distance calculation
$d_P(r,p) = 2-2 r^\top p$ we obtain {\em larger distance values}
compared to what would be obtained from a correctly normalized
prototype $d_P(r,p) = 2-2 r^\top \frac{p}{\|p\|}$ (since $\|p\|<1$).

As already mentioned, in order to have influence on the ranking of the
clusters (based on the prototype-distance), the prototype's norm must
differ within the clusters. This may easily happen if, for instance,
the amount of noise is not the same in all clusters: A set of linearly
increasing time series without any noise correlates perfectly, but by
adding random noise to the series the Pearson coefficient approaches 0
when the amplitude of the noise is continuosly increased. 

To demonstrate the relevance of these theoretical considerations we
construct an artificial test data set. It will consist of two clusters
(100 time series each) and a few additional time series serving as
probes that demonstrate the different behavior of the two algorithms.

\begin{figure}[htb]
\includegraphics[width=4cm,height=3cm]{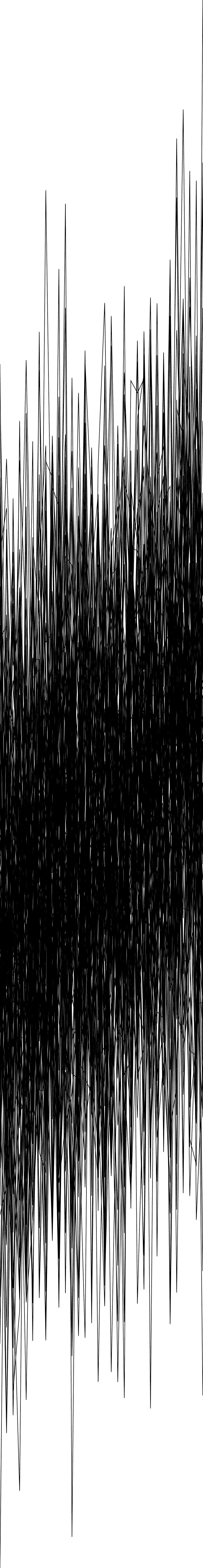}\hfill%
\includegraphics[width=4cm,height=3cm]{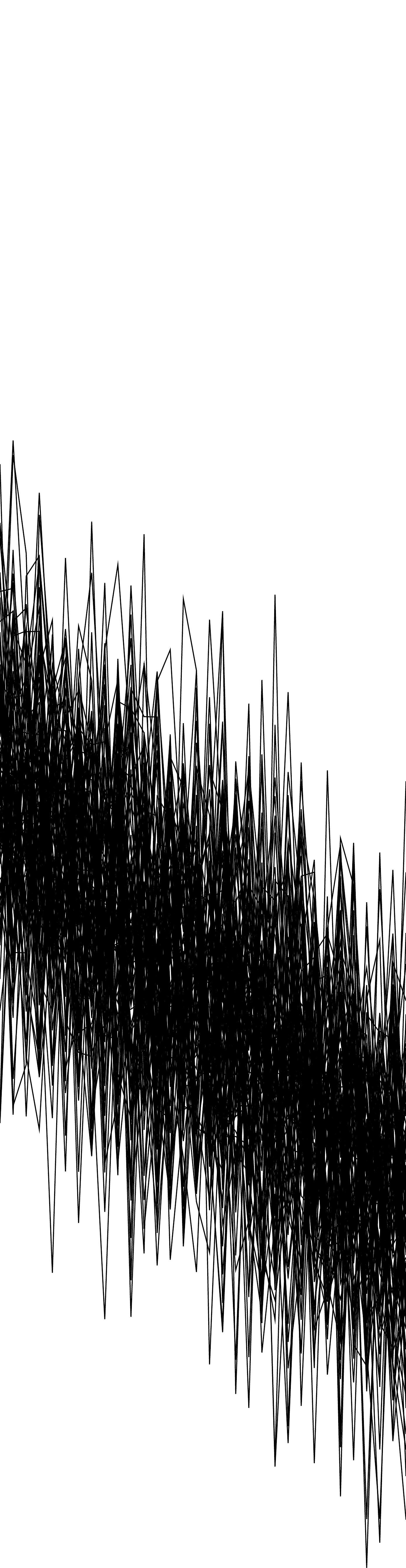}\hfill%
\includegraphics[width=4cm,height=3cm]{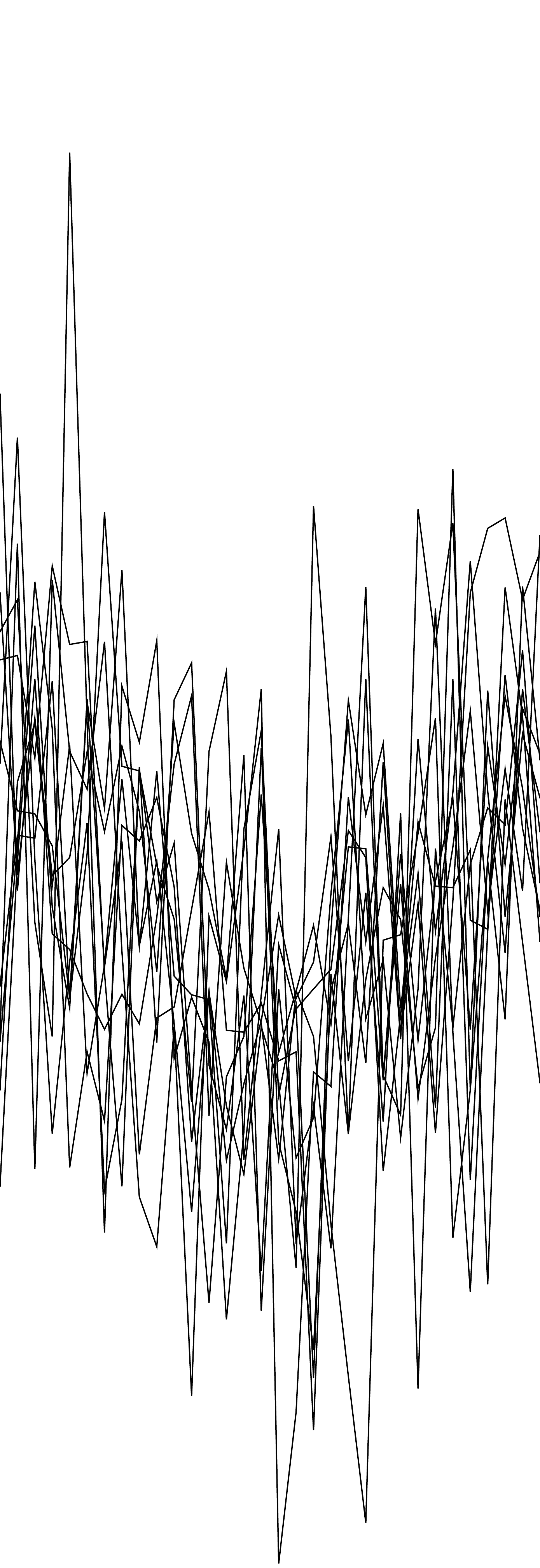}
\caption{\label{types:fig}Groups of time series in the test data set.}
\end{figure}

The test set contains a cluster of linearly increasing time series
$r_I=(0,1,2,..,29,30,31)$ as well as a cluster of linearly decreasing
time series $r_D=(31,30,29,...,2,1,0)$, 100 series per each
cluster. We add Gaussian noise with $\sigma=30$ and $\sigma=10$ to the
samples, that is, the data with increasing trend is more noisy than
the data with decreasing trend. Finally, we add another 10 time series
that equally belong to both clusters as they consist of a linearly
increasing and a linearly decreasing portion:
$r_M=(16,15,14,13,..2,1,0,1,2,..,13,14,15)$. The Pearson correlation
between $r_M$ and $r_I$ as well as $r_D$ is identical, therefore our
expectation is that -- after adding some Gaussian noise with
$\sigma=10$ to this group of time series -- on average half of them is
assigned to each cluster. For the experiment we want to concentrate on
how the small group of these ``probe series'' is distributed among both
clusters. The data from each group of time series is shown in figure
\ref{types:fig}.

To reduce the influence of random effects, we initialize the clusters
with linearly increasing and decreasing series (but the prototypes do
not stick to this initialization due to the noise in the
data). Clustering the z-score normalized data with standard k-Means
finally arrives at prototypes with a norm of $0.07$ (increasing
cluster) and $0.31$ (decreasing cluster) approximately. Instead of the
expected tie situation, all the probing series but one are assigned to
the cluster representing the increasing series (9:1). If we include
the normalization step for the Pearson-based k-Means, we obtain an
equal distribution among both clusters (5:5).

Given that the semantics of clustering z-score normalized time series
is so close to clustering via Pearson correlation, this example
clearly indicates that we may obtain counterintuitive partitions when
using standard k-Means. Clusters of poorly correlating data greedily
absorb more data than it would be justified by the Pearson
correlation. This undesired bias is removed by the properly modified
version of k-Means. Still, the chosen example is artificial and rather
drastic - the question remains how big the influence of this issue
really is in practice.

\subsection{Real World Datasets}

In order to demonstrate this on real world data sets, we use the time
series data sets for clustering tasks, available
online~\cite{Keogh:WEB:2006}. For each data set we use the number
of classes as the number of clusters\footnote{Obviously the number of
classes does not necessarily correspond to the number of clusters. But
for the sake of the experiments reported here, the exact match of the
true underlying clustering is not crucial, which is also the reason
why we do not make use of the available class (not cluster!)
information to judge the quality of the clustering methods considered
here.} and run each experiment five times with different random
selections of initial prototypes for each run of k-Means. During each
run we use the same vectors as initial prototypes for the different
types of algorithms to avoid problems with the unfamously instable
k-Means algorithm. In order to compare the result to the natural
instability of k-Means, we run a third experiment with a different
random initialization. That is, for each data set we do:
\begin{enumerate}
\item set the number of clusters to the number of classes,
\item normalize each time series (row) to have $\mu=0$ and $\sigma=1$,
\item create a clustering $\cal{C}_1$ using the classical k-Means
  algorithm with the Euclidean distance (no normalization),
\item create a clustering $\cal{C}_2$ using a k-Means algorithm using
  the Euclidean distance but prototype update equation (\ref{pupd:eq})
  with normalization after each iteration,
\item create a clustering $\cal{C}_3$ using a k-Means algorithm using
  the Euclidean distance and no normalization of the prototypes using
  a different (randomly chosen) initialization of the prototypes.
\end{enumerate}
We then compute the average entropy over five runs of the clustering
$\cal{C}_1$ with respect to $\cal{C}_2$ (called $E_{pear}$) and
$\cal{C}_1$ with respect to $\cal{C}_3$ (denoted by $E_{random}$).
The entropy measures are composed of the weighted average over the
entropy of all clusters in one clustering using the cluster indices of
the other clustering as class labels. $E_{pear}$ therefore measures
the difference between a clustering achieved with the correctly
modified k-Means algorithm and its lazy version where we only
normalize the input data but refrain from normalizing the
prototypes. $E_{random}$ shows the difference between two runs of
k-Means with the same setting but different initalizations to
illustrate the natural, internal instability of k-Means.  The
hypothesis would be that $E_{pear}$ should be zero or at least
substantially below $E_{random}$, that is, using the Pearson
correlation coefficient should not have a worse influence on k-Means
than different initializations.  Table~\ref{tab:exp} shows the
results. We show the name of the dataset and the number of clusters
in the first two columns. Length and number of time series in the
training data are listed in columns 3 and 4. The following four
columns show minimum resp.\ maximum value over our five experiments
for the two entropies.

\begin{table}[t]
\caption{Experimental results using different version of k-Means. See
  text for details.}
\label{tab:exp}
\centerline{
\begin{tabular}{|l|c|c|c||c|c||c|c|}\hline
Name              & \#clusters & \#length & \#size &
   \multicolumn{2}{c|}{$E_{pear}$} & \multicolumn{2}{c|}{$E_{random}$}  \\
                  &    &     &     &  min &  max &  min &  max \\\hline\hline
Synthetic Control &  6 &  60 & 300 & 0,32 & 0,36 & 0,41 & 0,62 \\ \hline
Gun Point         &  2 & 150 &  50 & 0,0  & 0,0  & 0,0  & 0,59 \\ \hline
CBF               &  3 & 128 &  30 & 0,0  & 0,0  & 0,30 & 0,85 \\ \hline
Face (All)        & 14 & 131 & 560 & 0,03 & 0,16 & 1,12 & 1,53 \\ \hline
OSU Leaf          &  6 & 427 & 200 & 0,06 & 0,26 & 0,93 & 1,25 \\ \hline
Swedish Leaf      & 15 & 128 & 500 & 0,0  & 0,03 & 0,83 & 1,09 \\ \hline
50Words           & 50 & 270 & 450 & 0,0  & 0,05 & 1,17 & 1,27 \\ \hline
Trace             &  4 & 275 & 100 & 0,0  & 0,0  & 0,50 & 0,67 \\ \hline
Two Patterns      &  4 & 128 &1000 & 0,03 & 0,11 & 0,86 & 1,41 \\ \hline
Wafer             &  2 & 152 &1000 & 0,0  & 0,0  & 0,0  & 0,0  \\ \hline
Face (Four)       &  4 & 350 &  24 & 0,0  & 0,0  & 0,62 & 1,19 \\ \hline
Lightning-2       &  2 & 637 &  60 & 0,0  & 0,0  & 0,0  & 0,71 \\ \hline
Lightning-7       &  7 & 319 &  70 & 0,0  & 0,07 & 0,35 & 0,99 \\ \hline
ECG               &  2 &  96 & 100 & 0,0  & 0,0  & 0,0  & 0,55 \\ \hline
Adiac             & 37 & 176 & 390 & 0,0  & 0,0  & 0,89 & 1,23 \\ \hline
Yoga              &  2 & 426 & 399 & 0,0  & 0,0  & 0,0  & 0,76 \\ \hline
\end{tabular}
}
\end{table}

As one can see clearly in this table, the difference between running
classical k-Means on z-score normalized input and the ``correct''
version with prototype-normalization is small. The clusterings are the
same ($E_{pear}=0$) when k-Means is reasonably stable
($E_{random}\approx 0$), that is an underlying reasonably well defined
clustering of the data exists (Gun Point and Wafer are good examples
for this effect). In case of an unstable k-Means, also the
normalization of the prototypes (or the lack of it) may change the
outcome slightly but will never have the same impact as a different
initialization of the same algorithm on the same data. Strong
evidence for this effect can be seen by
$\max\{E_{pear}\}\leq\min\{E_{random}\}$ in all case.  Note how in many
case even an unstable classic k-Means does not indicate any difference
between the prototype normalized and unnormalized versions,
however. Gun Point, CBF, Trace, and Face-Four among others are good
examples for this effect.

From these experiments it seems safe to conclude that one can simulate
k-Means based on Pearson correlation coefficients as similarity metric
by simply normalizing the input data without changing the normal
(e.g.\ Euclidean distance based) algorithm at all. It is, of course,
even easier to apply the discussed properties to classification or
clustering algorithms which do not need subsequent averaging steps,
such as k-Nearest Neighbor, hierarchical clustering, Kernel Methods
etc. In these cases a distance function using Pearson correlations can
be perfectly emulated by simply normalizing the input data
appropriately and then using the existing implementation based on the
Euclidean distance.

\section{Conclusion}

We have shown that the squared Euclidean distance on z-score
normalized vectors is equivalent to the inverse of the Pearson
Correlation coefficient up to a constant factor. This result is
especially of interest for a number of time series clustering
experiments where Euclidean distances are applied to normalized data
as it shows that the authors in fact were often using something close
to or equivalent to the Pearson correlation coefficient.

We have also experimentally demonstrated that the standard k-Means
clustering algorithm without proper normalization of the prototypes
still performs similar to the correct version, enabling the
use of standard k-Means implementations without modification to compute
clusterings using Pearson coefficients. Algorithms without ``internal''
computations, such as k-Nearest Neighbor or hierarchical clustering
can make use of the theoretical proof provided in this paper and be
applied without any further modifications, of course.

\bibliography{abr,mno,ts}
\bibliographystyle{splncs}

\appendix

\section{Update Equation for Pearson Distance}

The derivation with respect to the reference time series $p_{i}$ at time $t$ is
\[ \frac{\partial J}{\partial p_{i,t}} = 
     2 \lambda_i p_{i,t} - \sum_{j=1}^n u_{i,j} x_{i,t}  
\]
Setting the derivative to zero (necessary condition for minimum) yields
\begin{equation}\label{p:eq}
   p_{i,t} = \frac{1}{2\lambda_i} \sum_{j=1}^n u_{i,j} x_{j,t}
\end{equation}

Replacing $p_{i,t}$ in the normalization constraint (guaranteed by Lagrange multiplier) yields
\[
  1 = \|p_i\|^2 = \sum_{t=1}^T p_{i,t}^2 = \frac{1}{4\lambda_i^2} \sum_{t=1}^T \left( \sum_{j=1}^n u_{i,j} x_{j,t} \right)^2
\] 
or
\[
 \lambda_i = \frac{1}{2} \sqrt{ \sum_{t=1}^T \left( \sum_{j=1}^n u_{i,j} x_{j,t} \right)^2 }
\]

Inserting this expression into (\ref{p:eq}) gives us finally the following update equation
\[
p_{i,t} = \frac{1}{ \sqrt{ \sum_{t=1}^T \left( \sum_{j=1}^n u_{i,j} x_{j,t} \right)^2 } } \sum_{j=1}^n u_{i,j} x_{j,t}
\]

\end{document}